\documentclass[aps,prx,twocolumn,longbibliography,superscriptaddress,notitlepage,floatfix]{revtex4-2}

\usepackage{amsmath,amsfonts,amssymb,color,epsfig,graphics,graphicx,latexsym,theorem,url,multirow,diagbox}
\usepackage[colorlinks,colorlinks,citecolor=blue,linkcolor=blue,urlcolor=blue]{hyperref}
\usepackage[utf8]{inputenc}
\usepackage{booktabs}
\usepackage[T1]{fontenc}
\usepackage{courier}
\usepackage{listings}
\usepackage{dcolumn}
\usepackage{comment}
\usepackage{times}

\usepackage{algorithm}
\usepackage{algorithmic} 
\usepackage{braket}
\usepackage{bm}

\begin{document}
\title{Quantum Continual Learning Overcoming Catastrophic Forgetting}

\author{Wenjie Jiang}
\affiliation{Center for Quantum Information, IIIS, Tsinghua University, Beijing
100084, People\textquoteright s Republic of China}

\author{Zhide Lu}
\affiliation{Center for Quantum Information, IIIS, Tsinghua University, Beijing
100084, People\textquoteright s Republic of China}

\author{Dong-Ling Deng}
\email{dldeng@tsinghua.edu.cn}
\affiliation{Center for Quantum Information, IIIS, Tsinghua University, Beijing
100084, People\textquoteright s Republic of China}
\affiliation{Shanghai Qi Zhi Institute, 41th Floor, AI Tower, No. 701 Yunjin Road, Xuhui District, Shanghai 200232, China}

\begin{abstract}
Catastrophic forgetting describes the fact that machine learning models will likely forget the knowledge of previously learned tasks after the learning process of a new one. It is a vital problem in the continual learning scenario and recently has attracted tremendous concern across different communities. In this paper, we explore the catastrophic forgetting phenomena in the context of quantum machine learning. We find that, similar to those classical learning models based on neural networks, quantum learning systems likewise suffer from such forgetting problem in classification tasks emerging from various application scenes. We show that based on the local geometrical information in the loss function landscape of the trained model, a uniform strategy can be adapted to overcome the forgetting problem in the incremental learning setting. Our results uncover the catastrophic forgetting phenomena in quantum machine learning and offer a practical method to overcome this problem, which opens a new avenue for exploring potential quantum advantages towards continual learning.
\end{abstract}

\maketitle

\textit{Introduction.}\textemdash Humans and animals are able to incrementally acquire knowledge and skills from interacting experiences with the real world throughout their lifespan, which is functioned by a rich set of neurophysiological processes and biological mechanisms \cite{Murray2016Multisensory,Zenke2017Temporal}. This capability is a crucial reason why animals can survive the dynamically nondeterministic nature. Likewise, artificially-constructed computational systems may also be exposed to continuous streams of data and interactions and are desired to learn information from new experiences as well as preserving previously learned information \cite{Legg2007Universal,Legg2011Approximation,Hernandez-Orallo2010Measuring}. The ability to sequentially accumulate knowledge over time is referred as continual learning or lifelong learning \cite{Parisi2019Continual}. Continual learning is considered to integrate information from progressively non-stationary data where the number of tasks to be learned is not predefined \cite{Parisi2017Lifelong,Chen2018Lifelong}. Machine learning algorithms applied to a number of difficult problems have achieved enormous success \cite{Mnih2013Playing,Mnih2015Humanlevel,Silver2016Mastering,Silver2017Mastering,Krizhevsky2017Imagenet,Silver2018General} and they can achieve human-level performance or even outperform human-beings on many specific tasks such as playing Atari \cite{Mnih2013Playing,Mnih2015Humanlevel} and Go \cite{Silver2016Mastering,Silver2017Mastering}. Nevertheless, most machine learning algorithms are designed to capture the solution of a predefined problem and thus can hardly be reused when data of multiple tasks comes progressively. The main issue prevents those learning models from continual learning is the catastrophic forgetting \cite{McCloskey1989Catastrophic,Robins1995Catastrophic,French1999Catastrophic}, a fact that the performances on those earlier trained tasks abruptly decrease after learning new tasks because of the oblivion of those information gathered from previous data \cite{McCloskey1989Catastrophic}. Catastrophic forgetting is widely believed to be a crucial obstacle for achieving artificial general intelligence with neural networks  \cite{Goodfellow2015Empirical,Kemker2017Measuring}.

\begin{figure}[!h]
    \includegraphics[width=0.48\textwidth]{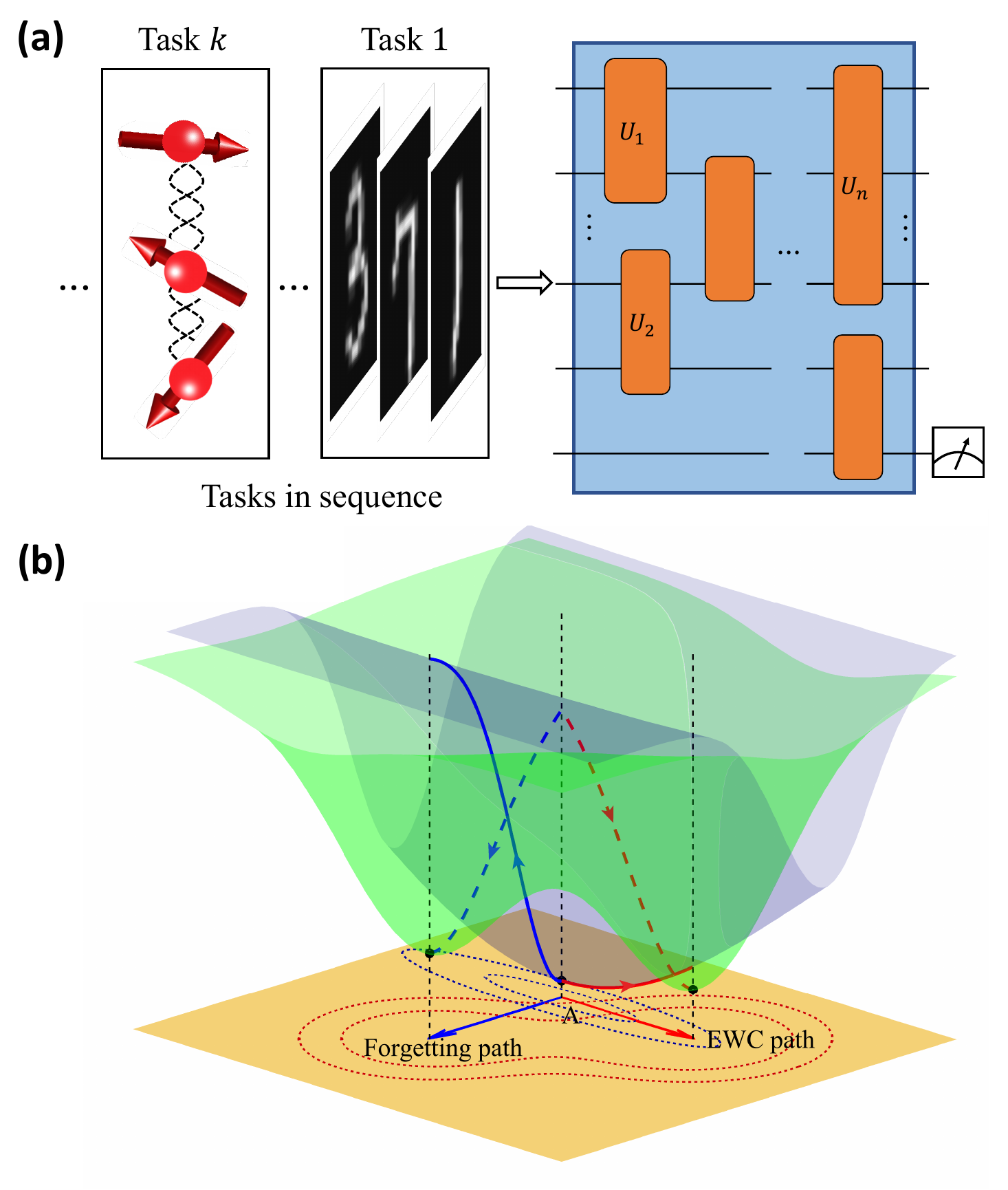}
    \caption{Illustration of quantum continual learning and the elastic weight consolidation (EWC) strategy adapted to overcome catastrophic forgetting. (a) Illustration of quantum continual learning. Here  different tasks are sequentially learned by the quantum classifier and the measurement result of a prefixed qubit denotes the output of this quantum classifier. (b) Geometric picture of the EWC method. The purple surface represents the loss landscape of task $1$ and the green surface represents the loss landscape of task $2$. After finding solution point $A$ for task $1$, retraining our quantum classifier for task $2$ using original training strategy leads to a significant increase of the loss for the previous task, which is shown as the forgetting (blue) path. In contrast, with the EWC method, retraining results in only a mild increase of the loss for task $1$ while achieving a low loss for the current task $2$, as depicted by the EWC (red) path.}\label{illu}
\end{figure}

In recent years, a number of variational quantum machine learning algorithms have been proposed to solve problems coming from the real world \cite{Lloyd2013Quantum,Lloyd2018Quantum, Amin2018Quantum, Cong2019Quantum}  and some of them have been demonstrated in proof-of-principle experiments to display their possible potential on practical applications \cite{Lamata2017Basic,Du2018Implementable,Hu2019Quantum,Saggio2021Experimental}. Those quantum learning algorithms exploit intrinsic properties underlying quantum computation systems like quantum superposition and quantum entanglement and promise exponential advantages compared with their classical counterparts \cite{Cong2016Quantum,Biamonte2017Quantum,Gao2018Quantum,Sarma2019Machine,Aaronson2015Read,
Carleo2019Machine,Liu2021Rigorous}. Despite those potential advantages and growing exciting results, there are still many unexplored aspects of quantum machine learning algorithms, which advocates for sustained research and development \cite{Alexeev2021Quantum,Awschalom2021Development,Altman2021Quantum}. In particular, similar to their classical counterparts, most quantum machine learning algorithms are meant to accomplish a predefined task and thus can hardly be generalized to learn multiple tasks sequentially. This prevents quantum learning agents from the ability of continual learning and remains a vital barrier to the accomplishment of quantum artificial general intelligence. In order to address this issue, it is crucial to investigate catastrophic forgetting in quantum machine learning models and to endow our learning agent based on quantum computer systems with the capacity of continual learning.

In this paper, we investigate the forgetting phenomena with a focus on a specific kind of learning models called quantum classifiers (see Fig. \ref{illu} for a pictorial illustration). We show that, similar to traditional classifiers based on classical neural networks, quantum classifiers based on variational quantum circuits likewise forget information after new parameter updating. Neuroscience suggests that the relations among different tasks can influence the forgetting degree in successive learning scenarios \cite{Zenke2017Temporal}. To determine whether our quantum classifiers are affected by the similarities among tasks, we examine two different types of relations between tasks: one in which the tasks are functionally identical but with permuted formats of the input and one in which the tasks are dissimilar in essential ways. Numerical experiments show that when trained on tasks one by one, both relations above can lead obvious decrease of the model's performances on those previously trained tasks. Furthermore, we interpret this problem more precisely and mitigate its influence via taking advantages of the local geometrical information in the loss function landscape of our adapted learning model. Based on our numerical results, we find that in certain situations, catastrophic forgetting in quantum machine learning can be overcome via the correct description of the local information and quantum continual learning is possible.

\begin{figure}[t]
    \centering
    \includegraphics[width=0.475\textwidth]{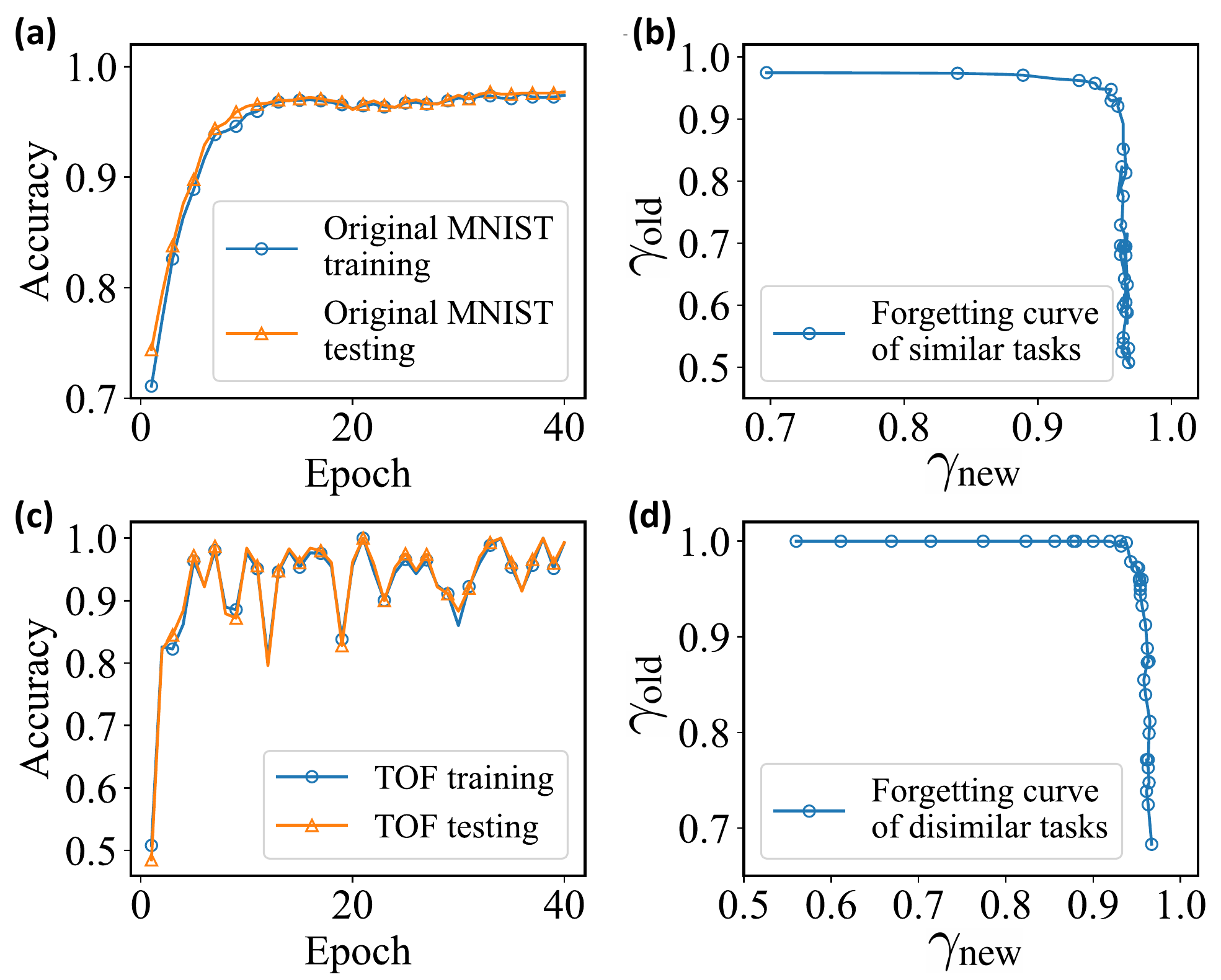}
    \caption{Catastrophic forgetting phenomena for quantum classifiers. (a) Learning curve for the original MNIST images. The blue line stands for the accuracy of the training set during the whole training process and the orange line for the accuracy of the testing set. (b) Forgetting curve for two tasks with large similarity. Here, $\gamma_{\text{old}}$ ($\gamma_{\text{new}}$) represents the accuracy of the quantum classifier on the old (new) task. (c) Learning curve of time-of-flight images for the task of classifying topological phases. (d) Forgetting curve of two dissimilar tasks:  the classifications of topological phases and hand-written digits.}\label{forgetting}
\end{figure}
\textit{Catastrophic forgetting in quantum learning.}\textemdash Quantum classifiers are a common set of variational quantum circuits targeted to accomplish classification tasks \cite{Dunjko2018Machine} and have been explored from many perspectives like vulnerability \cite{Lu2020Quantum,Liu2020Vulnerability,Gong2021Universal}, relation to feature space \cite{Schuld2019Quantum} and so on \cite{Grant2018Hierarchical,Blank2020Quantum,Du2020Quantum}. But there are also many aspects of quantum classifiers that remain unclear and catastrophic forgetting is a crucial one. The main purpose of classification tasks is to identify and sort a number of data into several different categories by recognizing and extracting meaningful fundamental features underlying data. Many practical problems can be abstracted as classification tasks, such as face recognization and object detection \cite{Russell2020Artificial}. 
To illustrate catastrophic forgetting phenomena in quantum machine learning, we use a pre-fixed variational quantum circuit to learn two classification tasks sequentially and observe the performance of our quantum classifier on the first classification task before and after the training process of the second task. In practice, many learning tasks share somewhat similar underlying structures \cite{Carroll2005Task} and can be naively transformed from one  to another. Learning English letters in capital and lower cases might be a gentle example of that. Abstractly, although they look not the same, letters in capital case and lower case share exactly the same rules and a simple encoding map can be adapted to transform from capital case to lower case and vice versa. For human-beings, the ability to identify and recognize letters in one format is useful when they are learning another format. However, instead of exploiting pre-learned information, quantum classifiers have the tendency to forget knowledge learned before.

To qualitatively understand this kind of forgetting, we directly construct a simple pair of tasks with similar underlying distributions and similar difficulties. One of those tasks is to classify hand-written digit images (0 and 9) randomly sampled from MNIST dataset \cite{LeCun1998MNIST}  and another task is to classify the same images with a pre-fixed pixel-permutation. Because the variational quantum circuit adapted here is generally ordinary and has no architecture exploiting graph information in images, we assume reasonably that a pre-fixed permutation of pixels does not change the original classification task significantly. We carry out extensive numerical simulations with a  classical computer \cite{Supp}.  The classification performances are plotted in Fig.~\ref{forgetting}{\color{blue}(a)} and Fig.~\ref{forgetting}{\color{blue}(b)}.  We firstly train our variational quantum classifier using original MNIST images and after 40 epochs of parameter updating, the accuracy of our classifier on target task is  high (larger than 95\%), which indicates that this circuit actually learns how to classify hand-written digit images. Then, we use the trained classifier to learn how to distinguish those permuted hand-written digit images. After several rounds of parameter updating, the accuracy of our classifier on the new task is also larger than 95\% even when their pixels are not aligned in normal order. Unfortunately, with the growth of the accuracy of our quantum classifier on this on-going new classification task, the accuracy on the previous task is getting worse and worse, indicating that the pre-learned information is being forgotten when our classifier is learning new information. After the training process with permuted images is finished, it is evident that information learned by the quantum classifier is almost refreshed and little knowledge from the previous task is left. This implies that catastrophic forgetting extends to quantum classifiers, even for learning tasks that share similar underlying structures.

To further illustrate the forgetting problem for quantum classifiers, we also consider the scenario of learning intrinsically dissimilar tasks. We choose two classification tasks emerging from disjoint areas in order to make sure they are not significantly related to each other. One of them is to classify topological phases of matter and the other remains as classifying MNIST hand-written digit images. Classifying different phases of matter is one of the central problems in modern condensed matter physics and many machine learning algorithms are proposed to deal with this task \cite{Dunjko2018Machine,Carleo2019Machine,Sarma2019Machine}, giving rise to a disciplinary research frontier connecting both machine learning and condensed matter physics. Here, we focus on classifying topological phases which are widely considered to be more challenging than that for the conventional symmetry-broken phases. We consider a two-dimensional (2D) square-lattice model for quantum anomalous Hall effect with following Hamiltonian \cite{Chang2013Experimental}:
\begin{eqnarray}
        H_{\mathrm{QAH}} &=& J_x \sum_{\mathbf{r}}(c_{\mathbf{r} \uparrow}^{\dagger} c_{\mathbf{r}+\hat{x} \downarrow}-c_{\mathbf{r} \uparrow}^{\dagger} c_{\mathbf{r}-\hat{x} \downarrow})+\mathrm{h.c.}  \\
        &+& i J_y  \sum_{\mathbf{r}}(c_{\mathbf{r} \uparrow}^{\dagger} c_{\mathbf{r}+\hat{y} \downarrow}-c_{\mathbf{r} \uparrow}^{\dagger} c_{\mathbf{r}-\hat{y} \downarrow})+\mathrm{h.c.} \nonumber \\
        &-& t \sum_{\langle\mathbf{r}, \mathbf{s}\rangle}(c_{\mathbf{r} \uparrow}^{\dagger} c_{\mathbf{s} \uparrow}-c_{\mathbf{r} \downarrow}^{\dagger} c_{\mathbf{s} \downarrow})+\mu \sum_{\mathbf{r}}(c_{\mathbf{r} \uparrow}^{\dagger} c_{\mathbf{r} \uparrow}-c_{\mathbf{r} \downarrow}^{\dagger} c_{\mathbf{r} \downarrow}), \nonumber
\end{eqnarray}
where $c_{\mathbf{r} \sigma}^{\dagger}\left(c_{\mathbf{r} \sigma}\right)$ is the fermionic creation (annihilation) operator with pseudo-spin $\sigma=(\uparrow, \downarrow)$ at site $\mathbf{r},$ and $\hat{x}, \hat{y}$ are unit lattice vectors along the $x, y$ directions. $J_x$  and $J_y$ are parameters characterizing the spin-orbit coupling strength along the $x$ and $y$ directions. The third and fourth terms describe the spin-conserved nearest-neighbor hopping (with strength $t$) and the on-site Zeeman interaction (with strength $\mu$), respectively. In the momentum space, this Hamiltonian has two Bloch bands and their topological properties can be diagnosed using the first Chern number:
\begin{equation}
    C_{1}=-\frac{1}{2 \pi} \int_{\mathrm{BZ}} d k_{x} d k_{y} F_{x y}(\mathbf{k}) \nonumber
\end{equation}
where $F_{x y}(\mathbf{k}) \equiv \partial_{k_{x}} A_{y}(\mathbf{k})-\partial_{k_{y}} A_{x}(\mathbf{k})$ is the Berry curvature with $A_{\nu}(\mathbf{k}) \equiv\left\langle\varphi(\mathbf{k})\left|i \partial_{k_{\nu}}\right| \varphi(\mathbf{k})\right\rangle$ being the Berry connection ($\nu=x, y$ and $|\varphi(\mathbf{k})\rangle$ is the Bloch wave function of the lower band), and the integration is  over the whole first Brillouin zone (BZ). In our numerical simulations \cite{Supp}, we first train a binary quantum classifier to assign labels of $C_1=0$ or $C_1=1$ to the time-of-flight images obtained from two distinct topological quantum states respectively and then use the trained classifier to learn how to identify hand-written digits. Our results are plotted in Fig.~\ref{forgetting}{\color{blue}(c)} and Fig.~\ref{forgetting}{\color{blue}(d)}, from which the catastrophic forgetting problem is manifested.    We observe that although the learning curve of time-of-flight images is not very smooth, the accuracy is reasonably high (larger than 95\%). However, during the learning process of classifying MNIST images, the accuracy for the previous task has an abrupt decrease with the growth of the accuracy for the new task.  At the end, the performance of the quantum classifier on learning topological phases becomes poor. 

\textit{Strategy overcoming catastrophic forgetting.}\textemdash Artificial general intelligence requires not only powerful representation architectures to produce complicated probability distributions, but also needs to preserve those experiences encountered before in order to imitate natural creatures. Quantum machine learning promises a potentially exponentially enlarged representation space to embed real-life distributions \cite{Lloyd2013Quantum,Biamonte2017Quantum,Dunjko2018Machine}. Nevertheless, results above show the undesirable fact that catastrophic forgetting phenomena occurs commonly in variational quantum classifiers and thus continual learning in the quantum machine learning domain cannot be gained for free. To overcome catastrophic forgetting is inevitable in the way towards quantum artificial general intelligence. In contrast to variational quantum classifiers illustrated above, humans are likely to remember knowledge and skills learned from previous experiences and take advantages from those related memories when coming across new environments. Recent research suggests that avoiding forgetting when animals are learning new tasks is related to the protection of some specific excitatory synapses strengthened by history experiences \cite{Yang2009Stably}. Adapting similar philosophy, we can presume that some parameters in variational quantum circuits are more important than others and should be protected carefully in following learning processes. This inspires a successful method overcoming forgetting phenomena in classical neural networks, i.e., the elastic weight consolidation method (EWC) \cite{Kirkpatrick2017Overcoming}. This is the intuition how we may overcome catastrophic forgetting in quantum classifiers as well.

From a high-level perspective, learning to assign different labels to data is a searching process in parameter space by updating parameters in the variational quantum circuit in order to optimize performance on training data. Usually, machine learning algorithm predefines a loss function assessing how good the performance of current parameters is,  and minimizes this loss function via some carefully-designed optimizers such as  stochastic gradient descent \cite{Bottou2004Stochastic} and adaptive moment estimation \cite{Kingma2017Adam}. This optimization procedure in the learning process is usually implemented on a high dimensional manifold and thus is highly nontrivial \cite{Goodfellow2016Deep}. Despite the difficulty for achieving the global minimum, finding a satisfactory local minimum in loss function landscape is likely accomplished in practice. Furthermore, for neural network based classifiers, there typically exist multiple local minimum points connected by simple curves \cite{Garipov2018Loss}, which form a substantially rich set of possible solutions for the target task. Thus, the aim of the training process is to achieve a satisfactory local minimum in the predefined loss function landscape.

Let us use the two-task (tasks $A$ and $B$) scenario as an example to discuss the EWC method adapted here to overcome the catastrophic forgetting problem in quantum machine learning. From the maximum likelihood estimation in statistical  learning \cite{Scott2002Maximum},  we should maximize the likelihood function $p(\bm{\theta}|\Sigma)$ of a model characterized by parameter $\bm{\theta}$ conditioned on the joint dataset $\Sigma=\Sigma_A+\Sigma_B$, where $\Sigma_A$ and $ \Sigma_B$ are datasets for the task $A$ and $B$, respectively. This likelihood function can be computed from the probabilities of given datasets by using the Bayes' rule under the assumption that tasks $A$ and $B$ are independent to  each other \cite{Supp}:
\begin{equation}
    \log p(\bm{\theta}|\Sigma)=\log p(\Sigma_B|\bm{\theta})+\log p(\bm{\theta}|\Sigma_A)-\log p(\Sigma_B). \nonumber
\end{equation}
Expanding the second term $\log p(\bm{\theta}|\Sigma_A)$ around the local minimum $\bm{\theta}^*_A$ for the task $A$, we have the following expression using Hessian matrix ${H}_{\bm{\theta}^*_A}$ with high order terms neglected:
\begin{equation}
        \log p(\bm{\theta}|\Sigma_A)=\log p(\bm{\theta}^*_A|\Sigma_A)+\frac{1}{2}(\bm{\theta}-\bm{\theta}^*_A)^T{H}_{\bm{\theta}^*_A}(\bm{\theta}-\bm{\theta}^*_A) \nonumber
\end{equation}
The Hessian matrix ${H}_{\bm{\theta}^*_A}$ is equal to the minus of the Fisher information matrix $F$ under specific regularity conditions \cite{Ly2017Tutorial}, which is an important concept in statistical learning  \cite{Kunstner2020Limitations,Ly2017Tutorial,Frieden1998Physics} and has been introduced to the quantum domain recently \cite{Petz2011Introduction,Liu2019Quantuma}. We thus approximate this posterior probability as a Gaussian distribution with the mean value given by $\bm{\theta}^*_A$ and the diagonal precision matrix given by the diagonal elements of $F$, and rewrite the loss function of the task $B$ as:
\begin{equation}
    \mathcal{L}(\bm{\theta}) = \mathcal{L}_{B}(\bm{\theta}) + \lambda\sum_i F_{i}\cdot(\theta_i-\theta_{A,i}^{*})^2. \nonumber
\end{equation}
Here, $\mathcal{L}_{B}(\bm{\theta})$ is the original loss function for the second task $B$, $F_i$ is the $i$-th diagonal element of the Fisher information matrix at the optimal point $\bm{\theta}^*_A$ for the previous task $A$,  and $\lambda$ is a hyper-parameter controlling the strength of this EWC restriction.  We refer to the Supplementary Material \cite{Supp} for more details. 

We can interpret this method more intuitively from a geometrical perspective, as illustrated in Fig. \ref{illu}{\color{blue}(b)}. The target of continual learning is to learn an adequate performance on the new task $B$ with no significant decrease of the performance on the previous task $A$. Based on this consideration, we can add a regularization to the original loss function when training on task $B$ to punish the deviation from the obtained optimal solution of task $A$ according to the importance of each parameter. To qualitatively evaluate the importance of different parameters in the quantum classifier, the Fisher information matrix of our trained quantum classifier is computed and its diagonal elements are used as weights of the penalties for the changes of different parameters. Under some mild regularity conditions, the Fisher information matrix characters the corresponding Hessian matrix, which describes the local curvature of the loss function landscape. Informally speaking, the gradient of each parameter nearly vanishes at the local minimum point in the loss function landscape of quantum classifiers, and the diagonal elements of corresponding Hessian matrix indicate the local curvatures along different directions at this local minimum point. Those curvatures explicitly suggest the significance of different parameters: a large curvature means the loss function value increases significantly even with a small shift of the corresponding parameter and a small curvature means the loss function value changes relatively mildly if the corresponding parameter is shifted a little bit. Thus, this regularity term forces those parameters in quantum classifier to update near the optimal solution of the previous task and punishes the alteration according to its local curvatures. We remark that this interpretation can help us extend this method to a more general scenario in a straightforward manner. In order to continually learn more than two tasks, we can simply compute the diagonal elements of the Fisher information matrix at the solution point of each task after finishing the corresponding training process and add a new regularization term according to those values to protect the quantum classifier's performance on the corresponding task when learning following tasks. As a result, the total loss will become very large if the current parameters are far away from those obtained optimal values for previous tasks. Consequently, minimizing the regulated loss function could achieve not only a decent accuracy for the new task, but also maintain a favorable performance on the previous ones. 
\begin{figure}[t]
    \centering
    \includegraphics[width=0.475\textwidth]{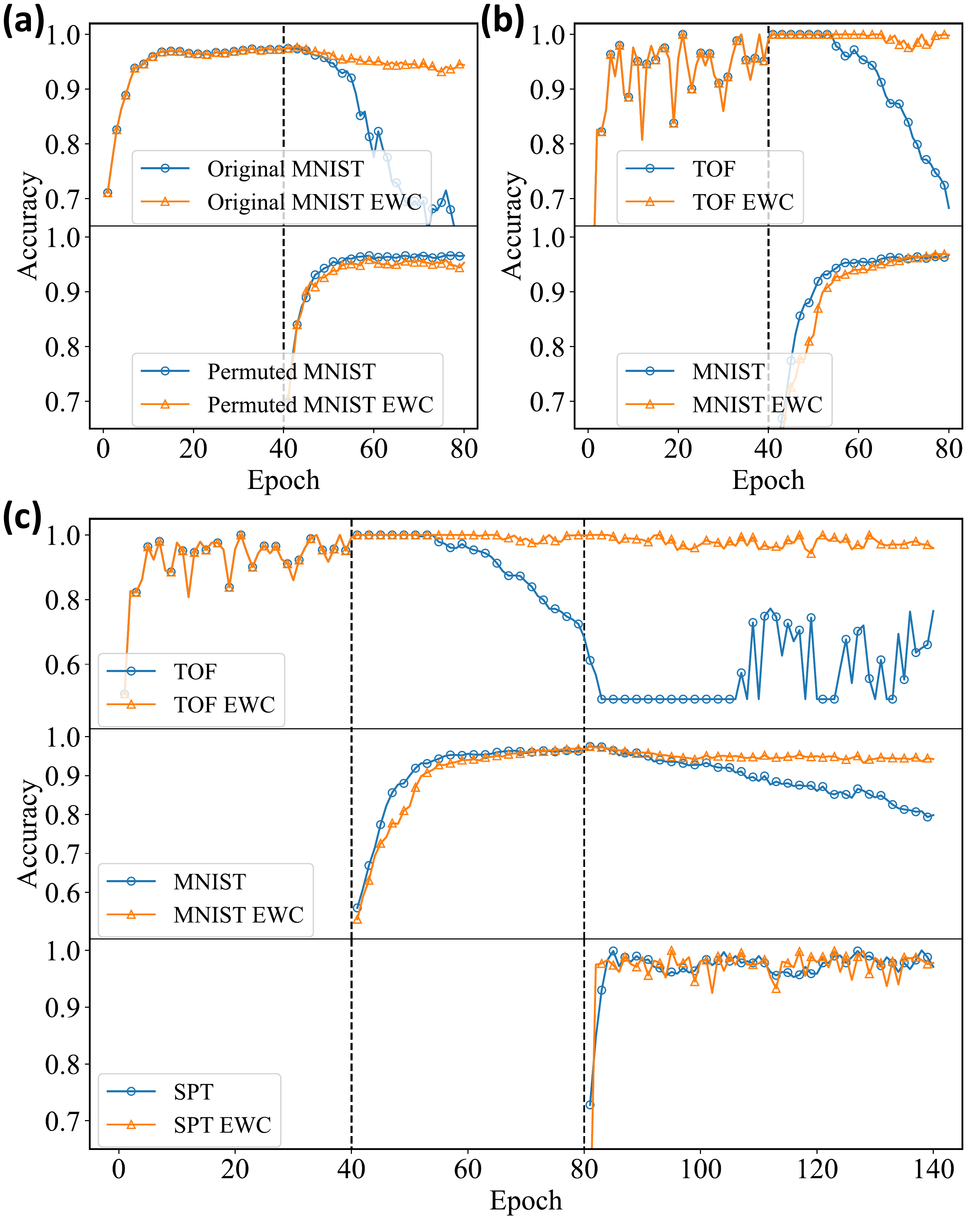}
    \caption{Performance benchmarking for the EWC strategy. (a) Learning curves of two similar tasks: classifying the original and pixel-permuted MNIST images. Blue lines plot the accuracies for the two tasks respectively trained without the EWC strategy, whereas orange lines show the corresponding results with using of the EWC strategy.  (b) Learning curves of two dissimilar tasks: classifying time-of-flight images from different topological phases and classifying the original MNIST hand-written images.  (c) Learning curves of three dissimilar tasks: classifying time-of-flight images, the original MNIST images, and the quantum states from different symmetry protected topological (SPT) phases. }\label{EWC}
\end{figure}
\textit{Numerical experiments.}\textemdash To benchmark the EWC method for quantum classifiers, we use the incremental learning setting above and adapt the similar learning settings on two pairs of classification tasks with different relations. The first pair of tasks is to classify MNIST digit images and their randomly pixel-permuted ones. We plot the continual learning result in Fig.~\ref{EWC}{\color{blue}(a)} and compare it with the forgetting result we discuss before. The upper panel of Fig.~\ref{EWC}{\color{blue}(a)} shows the full learning process of classifying original MNIST images with and without the EWC method respectively. During the second training phase targeting at classifying pixel-permuted MNIST images, the EWC method preserves the high accuracy of our quantum classifier on the previous task where evident performance reduction is avoided.   Meanwhile, the accuracy on the current task grows to the similar level as that of the quantum classifier trained without the regularization, as shown in the lower panel of Fig.~\ref{EWC}{\color{blue}(a)}. The second pair of tasks involves classifying time-of-flight images and hand-written digit images, as discussed above.  As for this pair of dissimilar classification tasks, our numerical results are plotted in Fig.~\ref{EWC}{\color{blue}(b)}, where an analogous performance-preserved behavior is clearly observed. 

We further test the EWC method on three dissimilar classification tasks emerging from different fields, and observe their learning curves in different training phases. In addition to classifying time-of-flight images and MNIST hand-written digits, we now add a third task---to classify the symmetry protected topological phases. We consider the following  Hamiltonian \cite{Smacchia2011Statistical}:
\begin{equation}
    H(h)=-\sum_i\sigma^x_{i}\sigma^z_{i+1}\sigma^x_{i+2}+h\sum_i\sigma^y_{i}\sigma^y_{i+1},
\end{equation}
where $\sigma^{x,y,z}_i$ are the usual Pauli matrices acting on the $i$-th spin and $h$ is a parameter describing the strength of the nearest-neighbor interaction. This Hamiltonian is exactly solvable and carries two well-studied quantum phases: one is the $\mathbb{Z}_2\times \mathbb{Z}_2$ symmetry protected phase characterized by a nonzero string order for $h<1$, and the other is an antiferromagnetic phase with long-range order for $h>1$. A quantum phase transition between these two phases occurs at $h=1$. Our results for learning three tasks sequentially with and without the EWC strategy are plotted in Fig.~\ref{EWC}{\color{blue}(c)}, from which the effectiveness of the EWC method is clearly manifested. Without adaption of the EWC strategy, the performance of the quantum classifier on classifying time-of-flight images and hand-written digits decreases notably as we train the classifier for the third task. In contrast, after the adaption of the EWC method the quantum classifier will maintain a reasonably good performance even at the end of the training process. We stress that those three tasks are coming from three distinct reach areas, and thus should share no significant underlying structure. Even in this situation, the proposed EWC strategy can still overcome catastrophic forgetting, which provides a possible way to achieve quantum continual learning in the future. 

\textit{Conclusion and outlook.}\textemdash In summary, we have investigated the catastrophic forgetting phenomena in the emergent interdisciplinary field of quantum machine learning. In particular, we showed that the catastrophic forgetting problem shows up commonly in quantum learning as well, and this problem could be overcome through the adaption of the EWC method. For concreteness, we carried out extensive numerical simulations involving a diverse spectrum of learning tasks, such as identifying real-life handwritten digit images, classifying time-of-flight images routinely obtained from cold-atom experiments,  and classifying quantum data for different symmetry protected topological phases. Our results not only reveal the notable catastrophic forgetting problem for quantum learning systems, but also propose an intriguing method based on Fisher information to overcome this problem. 

This work represents  only a preliminary step in the direction of quantum continual learning. Many important questions remain unexplored and deserve further investigations. First, in this work we have only considered the case of  supervised learning. How to extend our results to unsupervised and reinforcement learning remains unclear. We remark that such an extension might be highly nontrivial given the fact that in this scenario less or no priori knowledge will be available, or sometimes even the task boundaries are poorly defined \cite{Rao2019Continual}. Second, quantum machine learning holds the intriguing potential of exhibiting exponential advantages  \cite{Cong2016Quantum,Biamonte2017Quantum,Gao2018Quantum,Sarma2019Machine,
Aaronson2015Read,Carleo2019Machine, Liu2021Rigorous}. Yet, these advantages have only been explored in the context of a  predefined learning task. In the future, it would be interesting and important to investigate how to unambiguously demonstrate quantum advantages in the continual learning scenario.   In addition, recently quantum learning systems have been shown  to be notably vulnerable to carefully crafted adversarial examples and perturbations \cite{Lu2020Quantum,Liu2020Vulnerability,Gong2021Universal}. Along this line, it would be interesting and important to explore how quantum continual learning behaves under different adversarial settings. In particular, it would be of both theoretical and practical importance to study whether there exist universal perturbations that could deceive the quantum continual learning system for all the sequential tasks.  Finally, an experimental demonstration of quantum continual learning, especially with quantum advantages, should be a crucial step toward the long-term holy grail of achieving quantum artificial general intelligence. 

\begin{acknowledgments}
  We thank Peixin Shen and Weikang Li for helpful discussions.  This work is supported by the start-up fund from Tsinghua University (Grant. No. 53330300320), the National Natural Science Foundation of China (Grant. No. 12075128), and the Shanghai Qi Zhi Institute.
\end{acknowledgments}

\bibliography{QCLCF}

\clearpage
\onecolumngrid

\setcounter{secnumdepth}{3}

\makeatletter
\renewcommand{\thefigure}{S\@arabic\c@figure}
\renewcommand \theequation{S\@arabic\c@equation}
\renewcommand \thetable{S\@arabic\c@table}
\renewcommand \thealgorithm{S\@arabic\c@algorithm}

\begin{center} 
	{\large \bf Supplemental Material: Quantum Continual Learning Overcoming Catastrophic Forgetting}
\end{center}

\begin{figure*}[b]
    \includegraphics[width=\textwidth]{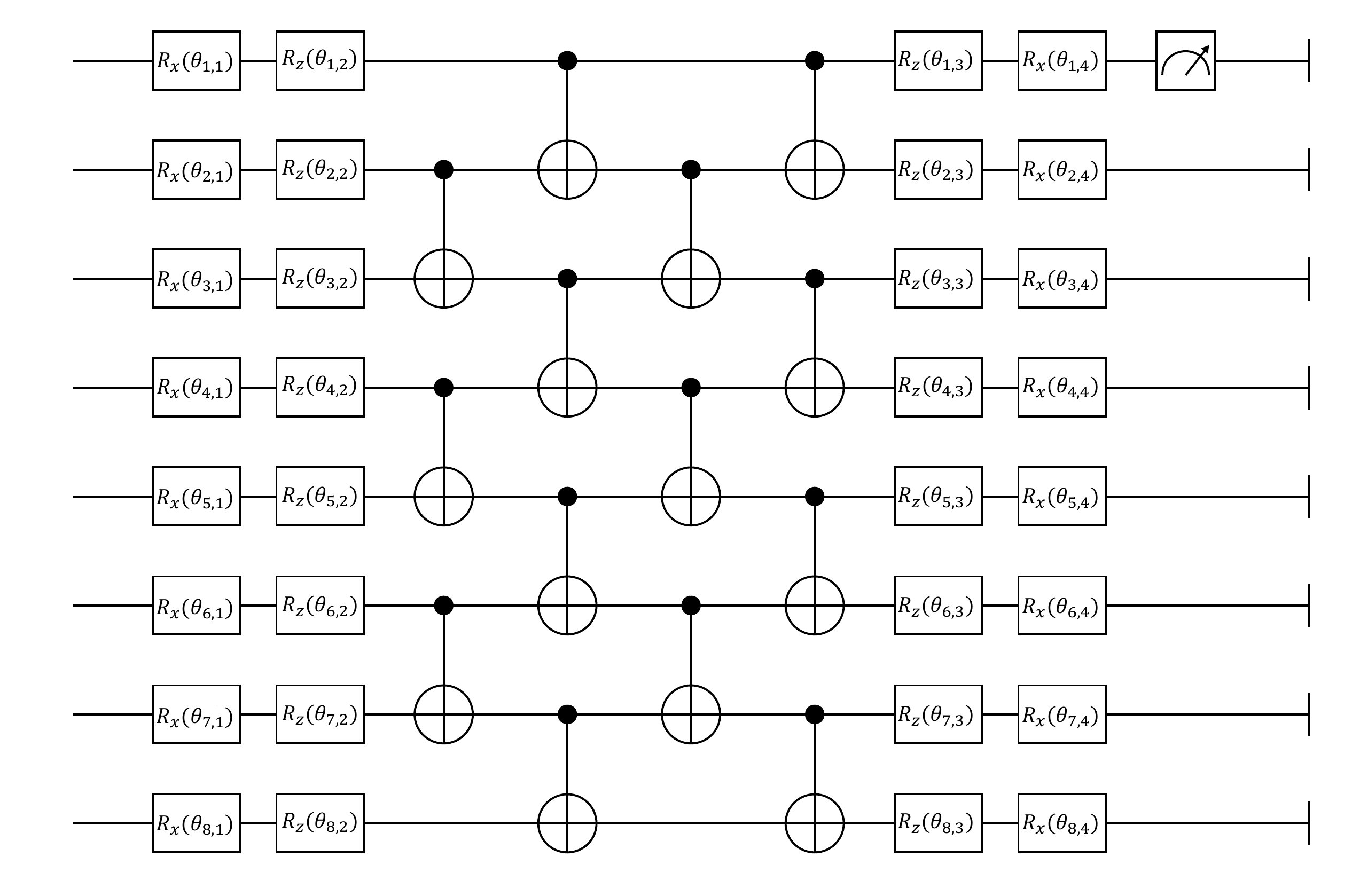}
    \caption{A single layer of the variational ansatz of our quantum classifier. This is a single layer of the ansatz. All single qubit gates in this ansatz are rotation gates ($R_x(\theta)=e^{i\theta/2X}$ and $R_z(\theta)=e^{i\theta/2Z}$). Those rotation angles $\theta_{i,j}$ ($i$ indicates the $i$-th qubit and $j$ indicates the $j$-th parameter of this qubit) in the rotation gates are the variational parameters of the quantum classifier and will be updated during the training process. Those CNOT gates are adapted here to introduce necessary entanglement among different qubits. We measure the first qubit and treat its output as the classification result of this quantum classifier. Our quantum classifier used in numerical simulations consists of ten repeated layers.}\label{classifier}
\end{figure*}

\section{The Setting}
\begin{figure*}[b]
    \includegraphics[width=\textwidth]{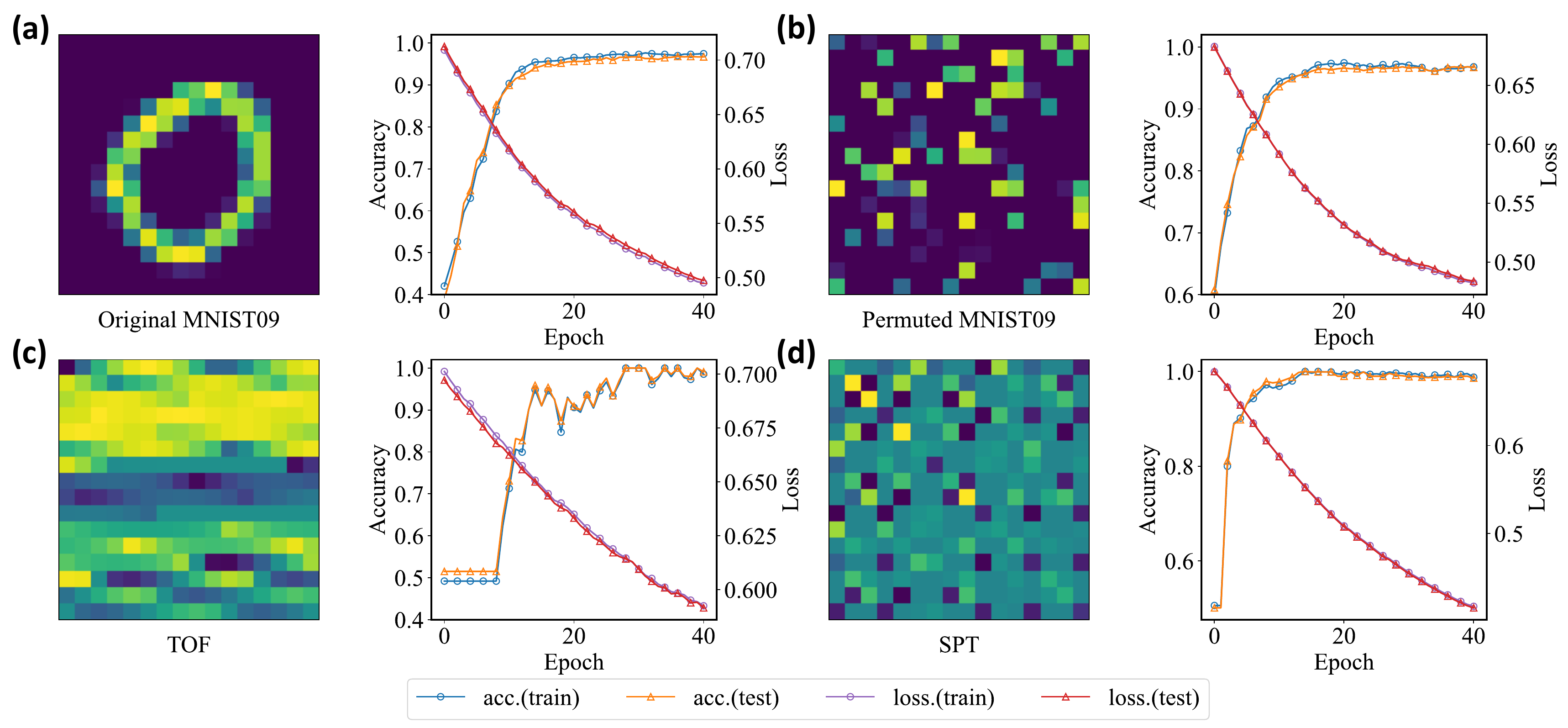}
    \caption{Results of learning single task. Here we show the classification performances of our quantum classifier on tasks used in our simulations of quantum continual learning and we also plot a sample image of each task: (a) the original MNIST image and its learning performance; (b) the permuted MNIST image and its learning performance; (c) the time-of-flight image and its learning performance; (d) the symmetry protected topological state and its learning performance.}\label{single}
\end{figure*}

Our numerical simulations are based on the open source package Yao.jl \cite{Luo2020yaojlextensible}. To illustrate the catastrophic forgetting phenomena, we randomly initialize an eight-qubit variational quantum circuit (as shown in Fig.~\ref{classifier}) as the ansatz for our quantum classifier, in which those rotation angles are variational parameters updated in the training process and unchanged in the inference process, and the CNOT gates is necessary to entangle all qubits since entanglement in quantum circuits is a key resource for potential quantum advantages. This variational architecture is hardware-efficient \cite{Kandala2017Hardwareefficient} and is capable to achieve satisfactory performances for our classification tasks (see Fig.~\ref{single}). Besides, this architecture does not take advantages of the specific structure information of datasets.

All data encountered in our numerical simulations consists of 256 features and can be represented by eight qubits using amplitude encoding. For the original MNIST hand-written digit images, those $28\times 28$-pixel images \cite{Lecun2010Mnist} are reduced to $16\times 16$-pixel images (see Fig.~\ref{single}{\color{blue}(a)}), so that we can simulate this quantum learning process with moderate classical computational resources. Then, we randomly choose a permutation of the 256 pixels and apply it for all images, which produces a new dataset consisting of pixel-permuted images (see Fig.~\ref{single}{\color{blue}(b)}). For time-of-flight (TOF) images, we diagonalize the Hamiltonian of quantum anomalous Hall effect with an open boundary condition and calculate the atomic density distributions with different spin bases for the lower band in momentum space to obtain input data. We vary the strength of the spin-orbit coupling and the strength of the on-site Zeeman interaction in both the topological and topologically trivial regions to generate several thousand data samples (see Fig.~\ref{single}{\color{blue}(c)}). For the symmetry protected topological state (SPT), we consider the model involving eight spins and exactly diagonalize its Hamiltonian to obtain the ground state which can be naturally represented using eight qubits (see Fig.~\ref{single}{\color{blue}(d)}). In this work, we use amplitude encoding to convert the data of our classification tasks into the input quantum states for the quantum classifier. 

The process of sequential learning is divided into different phases and our quantum classifier are trained with only one specific dataset in each training phase. For example, to illustrate the catastrophic forgetting phenomena, we first use the randomly initialized quantum classifier to learn to classify original MNIST images. After a satisfactory performance is obtained, this classifier are trained to distinguish permuted MINIST images. The results of different learning phases are shown in the main text, where the forgetting phenomena is revealed. As for continual learning via EWC method, the Fisher information matrix for each task is computed after the corresponding training phases and is stored for those following training phases.

\section{Elastic Weight Consolidation}

From a high-level perspective, overcoming catastrophic forgetting in quantum continual learning requires protecting the learned knowledge of those previous tasks, as well as learning the new-coming knowledge of following tasks \cite{Rao2019Continual,Aljundi2019Continual}. So our quantum learning model should have enough capacity to store those information. Besides, appropriate management of model's capacity is required to achieve quantum continual learning in practice. EWC method offers a practical method to do the capacity management: it estimates the necessary capacity for previous tasks and refreshes the rest part which contains rare information about those previously trained tasks. To do this, EWC method evaluates the importance of each variational parameter in the quantum classifier and only allows significant twist for those relatively unimportant ones. 

We then give a detailed mathematical derivation of EWC method. For simplicity, we concern the two-task scenario here and use the similar philosophy to explicitly write down the result for the multi-task scenario. From the perspective of maximum likelihood estimation \cite{Scott2002Maximum}, we explore all possibilities of parameters $\bm{\theta}$ of the quantum classifier to maximize the likelihood function $p(\bm{\theta}|\Sigma)$, where $\Sigma=\Sigma_A+\Sigma_B$ is the total dataset ($\Sigma_A$ and $\Sigma_B$ are datasets for task $A$ and task $B$ respectively and we assume that these two tasks are independent to each other). So we have expression
\begin{align*}
    \log{p(\bm{\theta}|\Sigma)}&=\log{\left(\frac{p(\Sigma_B|\Sigma_A,\bm{\theta})p(\Sigma_A,\bm{\theta})}{p(\Sigma_A,\Sigma_B)}\right)}=\log{\left(p(\Sigma_B|\bm{\theta})\cdot \frac{p(\bm{\theta},\Sigma_A)}{p(\Sigma_A)} \cdot \frac{1}{p(\Sigma_B)}\right)}\\
    &=\log{p(\Sigma_B|\bm{\theta})}+\log{p(\bm{\theta}|\Sigma_A)}-\log{p(\Sigma_B)},
\end{align*}
where the first and third equation use the Bayes' rule and the second equation uses the independence condition. As shown in the main text, we have Taylor Series for the second term: 
\begin{equation}
    \log p(\bm{\theta}|\Sigma_A)=\log p(\bm{\theta}^*_A|\Sigma_A)+
    \frac{1}{2}(\bm{\theta}-\bm{\theta}^*_A)^T{H}_{\bm{\theta}^*_A}(\bm{\theta}-\bm{\theta}^*_A)\nonumber.
\end{equation}
It is worthwhile to mention that from the perspective of parameter estimation \cite{Wiley2007Precise}, this treatment means that we sample parameters from a multivariate normal distribution:
\begin{equation}
    p(\bm{\theta}|\Sigma_A)\propto \mathcal{N}\left(\bm{\theta}^*_A, {H}_{\bm{\theta}^*_A}^{-1}\right),
\end{equation}
where the optimal solution $\boldsymbol{\theta}_{A}^{*}$ for task $A$ is the mean value of this normal distribution and $H_{\boldsymbol{\theta}_{A}^{*}}^{-1}$ is the precision matrix  ($\left(H_{\bm{\theta}_{A}^{*}}\right)_{ij}=\frac{\partial}{\partial\theta_i\partial\theta_j}\log p(\bm{\theta}|\Sigma_A)|_{\bm{\theta}^*_A}$ is the Hessian matrix at the optimal solution $\bm{\theta}^*_A$ for task $A$ and is equal to the minus of the Fisher information matrix $F$ under some specific conditions \cite{Ly2017Tutorial}). We can rewrite the quadratic term using the Fisher information matrix and absorb it into the likelihood function of sequential tasks. This leads to the loss function for the second task in our scenario:
\begin{equation}
    \mathcal{L}(\boldsymbol{\theta})=\mathcal{L}_{B}(\boldsymbol{\theta})+\lambda\cdot (\bm{\theta}-\bm{\theta}^*_A)^T{F}_{\bm{\theta}^*_A}(\bm{\theta}-\bm{\theta}^*_A)\nonumber.
\end{equation}
To reduce the potential storage and computation overhead for those possible large quantum models, we use the diagonal elements of the Fisher matrix as the weights of variational parameters and neglect those off-diagonal entries, which will be discussed later. Thus, we could add the regularization term shown in the main text to the loss function of the second task in order to maximize the likelihood function of joint tasks. 

\begin{figure}[b]
    \includegraphics[width=0.7\textwidth]{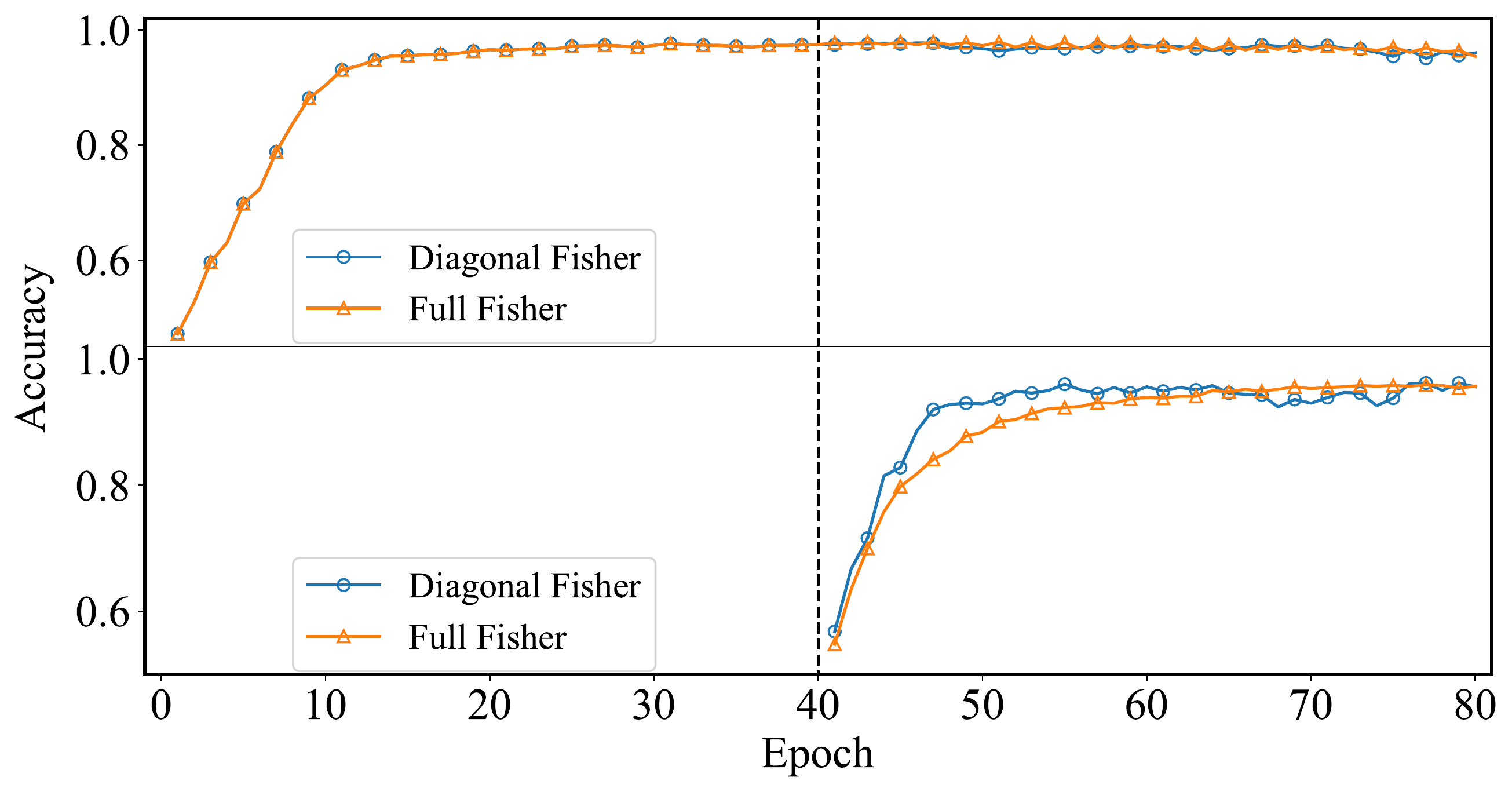}
    \caption{Comparison between the learning result of using diagonal elements of the Fisher matrix and that of using the full Fisher matrix. We train our quantum classifier using the original MNIST dataset and then adapt two kinds of regularization terms to train this classifier using the new-coming permuted MNIST dataset. The learning settings for both cases are exactly the same except the strength parameter $\lambda$.}\label{fisher}
\end{figure}

For continual learning of more than two tasks, we can compute the regularization term for each trained task and add them together to overcome catastrophic forgetting:
\begin{align*}
    \mathcal{L}(\bm{\theta})=&\mathcal{L}_{0}(\bm{\theta})+\lambda_A \sum_{i} F_{i}^{(A)} \cdot\left(\theta_{i}-\theta_{A, i}^{*}\right)^{2} + \lambda_{A,B} \sum_{i} F_{i}^{(A,B)} \cdot\left(\theta_{i}-\theta_{A, B, i}^{*}\right)^{2}+...,
\end{align*}
where $\mathcal{L}_{0}(\boldsymbol{\theta})$ is the original loss function for current task given current parameters $\boldsymbol{\theta}, F_{i}^{(A)}$ is the $i$-th diagonal element of the Fisher information matrix at the optimal point $\boldsymbol{\theta}_{A}^{*}$ for previous task $A$, $\lambda_A$ is a hyper-parameter controlling the strength of this EWC restriction and so on.

\section{Reasons for neglecting off-diagonal elements}
In our numerical simulations, the quantum classifier consists of 248 variational parameters, in which computing and storing the full Fisher matrix is not very hard. Nevertheless, if the number of parameters gets larger and larger to match the exponentially growing dimensionality of the Hilbert space, computing and storing its full Fisher matrix can be quite challenging. From a more practical perspective, we use the diagonal elements of the Fisher matrix which can be estimated by the first order derivative \cite{Chi2006Fundamentals}.

To compare the learning result of using the diagonal elements of the Fisher matrix and that of using the full Fisher matrix, we train our quantum classifier using the original MNIST images and the permuted MNIST images sequentially. In this simulation, the diagonal elements of the Fisher matrix and the full Fisher matrix are adapted as the metric to quantify the derivative distance in the parameter space respectively. The results in Fig.~\ref{fisher} shows that the performances of both metric choices are at the same level. We remark here that in consideration of the summation of those off-diagonal elements, we manually lower down the strength parameter $\lambda$ in the simulation of using the full Fisher matrix. The similar performances between those two learning scenarios indicate that neglecting those off-diagonal elements in the Fisher matrix has no significant influence on the results of quantum continual learning. Thus, we use the diagonal elements as our distance metric in all other numerical simulations.

\section{More numerical results}

In this section, we give more results of quantum continual learning. Performances of learning single tasks are shown in Fig.~\ref{single} and one sample image of each dataset is plotted. Those results indicate that our quantum classifier is capable to achieve satisfactory performances on those chosen classification tasks.

\begin{figure}[t]
    \includegraphics[width=0.5\textwidth]{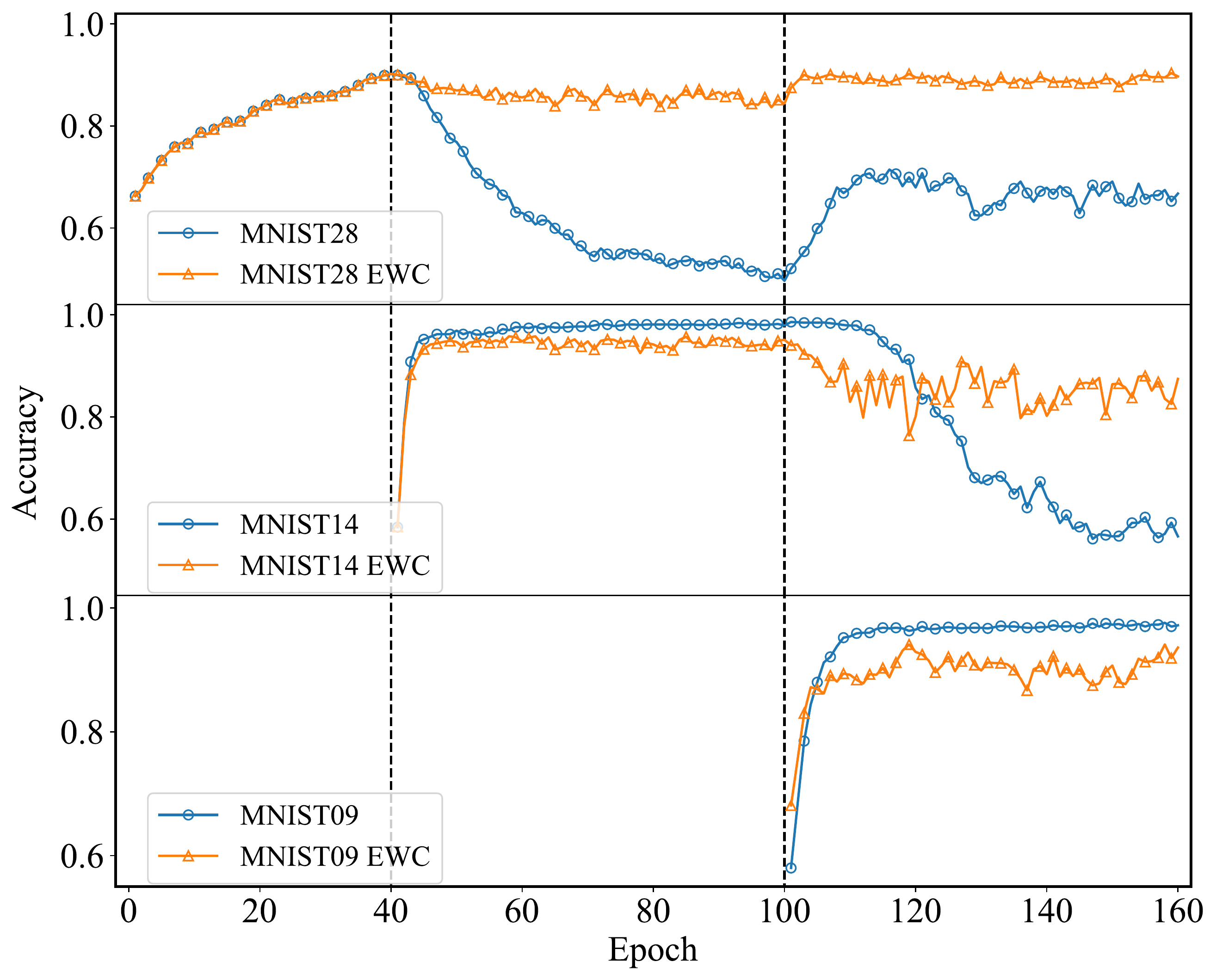}
    \caption{Illustration of quantum continual learning of classifying different MNIST images. Learning curves of three related tasks: classifying digit 2 and digit 8, classifying digit 1 and digit 4, and classifying digit 0 and digit 9.}\label{res}
\end{figure}

In the main text, we show that quantum continual learning of two-task case can be accomplished when those two problems are similar or dissimilar to each other. As a complementary example, we also simulate the quantum continual learning of two related problems. We use MNIST images of different digits to construct several classification tasks and find that the continual learning of this kind of tasks can also be accomplished (see Fig.~\ref{res}).

We group MNIST hand-written images of different digits to construct several binary classification tasks and use them to train our quantum classifier. For multi-task cases, we choose three pairs of digits and use our quantum classifier to classify their hand-written images. We first train our quantum classifier using images of digit 2 and images of digit 8, which ends with a high classification performance ($>90\%$). Then, we train this quantum classifier to identify digit 1 and digit 4. In the favor of EWC method, our quantum classifier behaves reasonably well at both tasks after the second training phase. Sequentially, we train this circuit to classify digit 0 and digit 9, and find that our quantum classifier can perform relatively well in all three different classification tasks after those training processes.

We also notice that in the continual learning scenario, the performance of our quantum classifier on each task has a slight reduction compared with that in the single task learning scenario. Intuitively, this is caused by an inevitable small deviation from the optimal solution of a single task to the optimal solution of the joint task.

\end{document}